\documentclass[runningheads]{llncs}

\usepackage[hidelinks]{hyperref}
\usepackage{graphicx}
\usepackage{subfigure}
\usepackage[utf8]{inputenc}

\begin{document}

\title{Which Country is This? Automatic Country Ranking of Street View Photos\thanks{The bulk of this work was undertaken by the first author during his Master's thesis. We gratefully acknowledge the funding by the \emph{High Tech Agenda Bavaria} (\url{https://www.bayern.de/wp-content/uploads/2019/10/Regierungserklaerung_101019_engl.pdf}) to the third author that supported part of this research.}}

\author{Tim Menzner\inst{1} \and Florian Mittag\inst{1}\orcidID{0000-0002-6472-0221} \and Jochen L.~Leidner\inst{1,2}\orcidID{0000-0002-1219-4696}}

\institute{
  Coburg University of Applied Sciences and Arts,\\Friedrich-Streib-Str. 2, 96450 Coburg, Germany.
  \and 
  University of Sheffield, Regents Court, 211 Portobello, Sheffield S1 4DP, UK\\
  Corresponding author contact: \email{tim.menzner@hs-coburg.de}
}

\authorrunning{T.~Menzner, F.~Mittag and J.~L.~Leidner}
\titlerunning{Which Country is This?}

\maketitle

\begin{abstract}
  In this demonstration, we present Country Guesser, a live system that guesses the country that a photo is taken in.
  In particular, given a Google Street View image, our federated ranking model uses a combination of computer vision, machine learning and text retrieval methods to compute a ranking of likely countries of the location shown in a given image from Street View.
  Interestingly, using text-based features to probe large pre-trained language models can assist to provide cross-modal supervision.
  We are not aware of previous country guessing systems informed by visual and textual features.
\end{abstract}

\section{Introduction}

If someone looked at a physical or electronic photo, one of the natural
questions one may ask is ``Where was this taken?''. For instance, the
Geographical Magazine of the Royal Geographical Society features a monthly
competition showing a photo and asking its readers to identify the place depicted \cite{RGS:2022;Geogr}.

In this demonstration, we
present a new system that addresses an easier sub-problem, namely: given a
Google Street View image in specific, which \emph{country} is shown on the image?
Our approach is to use a set of individual classifiers, such that the
individual signal evidence is combined to create a ranked list of
countries. Interestingly, both visual and
textual features turn out to be helpful in the task.

\section{Related Work} 

One early notable contribution for solving the problem of determining the geolocation of images was the winning entry of the ``where am I'' contest carried out during the 2005 International Conference on Computer Vision. This approach featured a huge database of city street scenes tagged with GPS locations and used SIFT to find correspondences \cite{zhang2006image}:
classical feature matching techniques were combined with landscape classification the classical feature matching by IM2GPS, which was, according to its authors, ``the first to be able to extract geographic information from a single image'' \cite{hays2008im2gps}.
Later, the topic was addressed in a wide range of papers (see
\cite{brejcha2017state} for a detailed survey).
One of the first papers that used machine learning for geolocating images was PlaNet \cite{weyand2016planet}, which divided earth's surface into thousands of grid cells. Then, 126 million images with geo information were assigned to the respective cells. A convolutional neuronal network was trained with these images to output a probability score for each and every cell.
Zamir et al. \cite{zamir2010accurate} focused specifically on
Google Street View images. They used  trees of indexed SIFT descriptors and
100,000 Street View images from Pittsburgh, PA and Orlando, FL to find GPS coordinates
for (not necessarily Street View) pictures from these cities. 
Also using Street View and inspired by the game GeoGuesser, \cite{Nirvan2020} 
combined Google Street View imagery and machine learning to make educated
guesses about photos from Street View’s coverage of the United States.
In contrast, our system provides a ranked list of countries and has
world wide geographic scope.\footnote{Between acceptance of this paper and our
preparation of the publication version, we found another work, \cite{Alamayreh-etal:2022:ArXiv} which attempts country guessing of photos that are not necessarily
from Street View, which is very much in the spirit of this paper.
}

\section{Graphical User Interface}

\begin{figure}
    \centering
    \includegraphics[width=.9\textwidth]{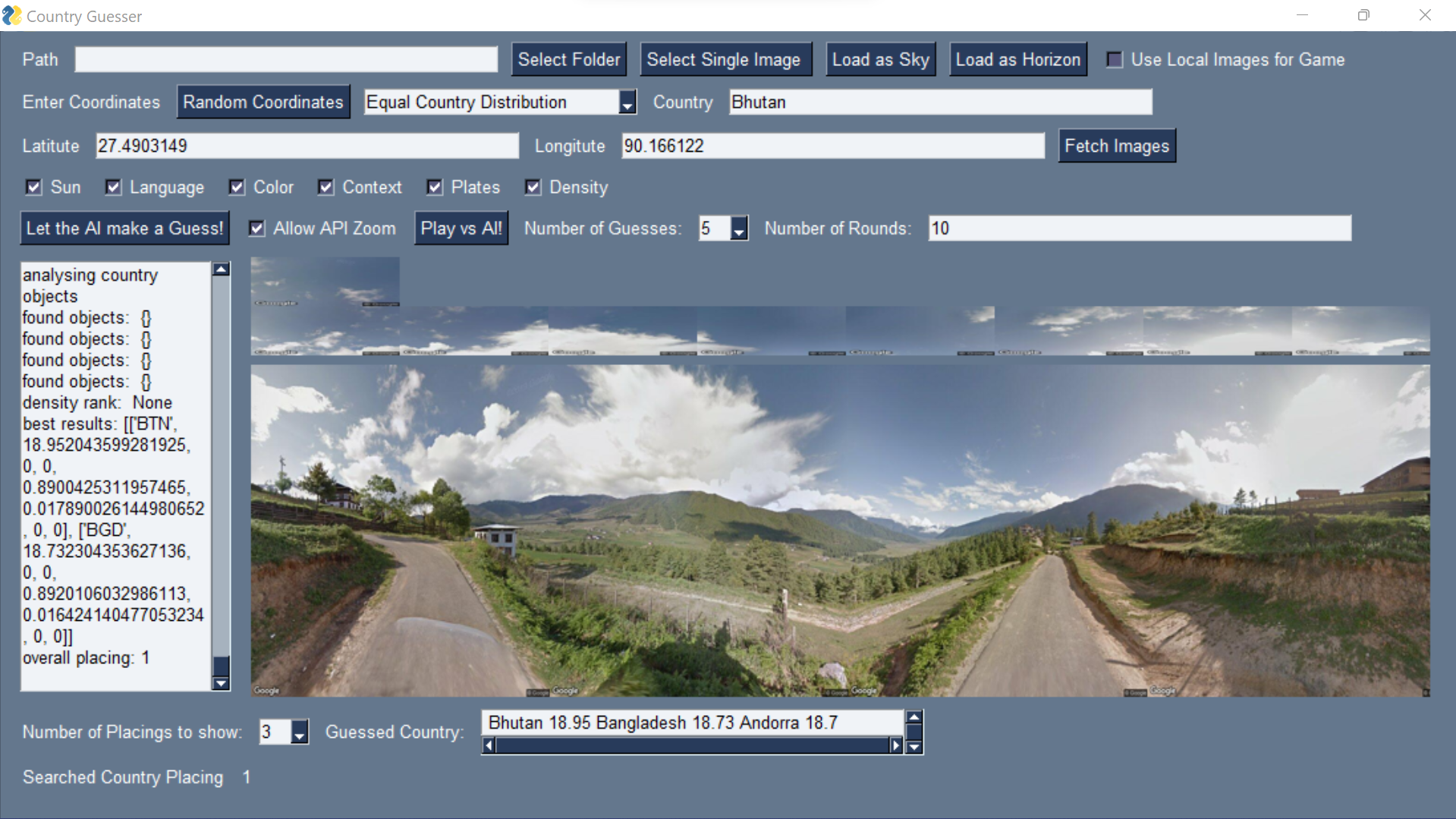}
    \caption{Country Guesser recognizes a Google Street View image as located in Bhutan.}
    \label{fig:gui}
\end{figure}

The system's graphical user interface is shown in Figure \ref{fig:gui}. The system can operate in two basic modes: in the first mode,
users can load a photo either from the local hard drive or pulled via Google Street View's online API for an automated guess, while additional information about the results of each individual module are provided. There is also the possibility to switch into a game mode, where the user can directly compete against the system.

\section{Method and Implementation}

\subsubsection{External Data.}
We used Geo-JSON country boundary polygons from the Natural Earth, public domain dataset.
We identify countries based on cues, a database maintains this evidence, such as what language is spoken where; information about the language of text in an image is not much of use without such background knowledge. Therefore, every featured country also gets a fact sheet in the form of an external JSON file. The information for this fact sheet was collected manually.
Overall, 110 countries were found to have sufficient Street View coverage to be included in this project.\\
\textbf{Solar Position.}
We narrow down the list of candidate countries, leveraging the sun's position via celestial direction. 
As the earth orbits the sun with an axial tilt of about 23.4 degrees, the relative position of the sun throughout the day is different between the two hemispheres. With a southern sun position only possible on the Northern Hemisphere and northern sun position only possible on the Southern Hemisphere (except for the tropics), this concept can be used to determine the hemisphere where the panorama image was taken.
For detecting the sun position, our approach is to split the panorama, which covers a 360 degrees field of view, into separate images pitched towards the sky, to search for the brightest image.\\ 
\textbf{Text and Languages.}
Street View, as the name suggests, mainly consist of images taken from streets, so often street signs, billboards etc. showing some kind of text are visible. 
We combine optical character recognition (OCR) using \emph{EasyOCR} \cite{Kittinaradorn2022easyocr} to obtain text before checking the most likely language using the Python \emph{langdetect} \cite{nakatani2010langdetect} and \emph{lingua} \cite{stahl2022lingua} libraries and, finally, we scan text for place names mentioned and knowledge of the country the place is located in \cite{Leidner:2008}.\\
%
\textbf{Coloration.}
Differences in climate and vegetation lead to different colors being associated with the ``look'' of a country: Ireland for example is well known to be very green.
To objectively model this relationship between color palette and country, we use one histogram for each RGB channel.
The average histograms for a country can be generated by looking at the histograms of a number of images from that country, summing up the total number of pixels for each intensity value and respectively dividing it by the number of images used. 
The average differences across all positions for all histogram pairs are then used to calculate an overall similarity score and to rank all countries accordingly.\\
\textbf{Captions.}
A powerful feature was the descriptive text provided by automatic caption generation (we used the \emph{ClipCap} model \cite{mokady2021clipcap}).
One more concept that proved to be useful was analysing differences in the frequency of individual words being featured in the image descriptions generated by image captioning models.
We generated average ``word lists'' for each country as an offline step. These lists contain a  value indicating the average occurrence of all words that were used by the model when describing an image from the country during its generation (with non-descriptive words like articles and adverbs being filtered out).\\
\textbf{Car License Plates}.
Despite license plates being blurred on Street View imagery for privacy reasons, their colours are still visible and can be used to guess the country.\\
\textbf{Objects.}
We use  \emph{YOLO} \cite{Jocher:2022} to generate "average object lists" and use them to guess the country, in the same manner as with the previously described word lists.

\begin{figure}
    \centering
    \includegraphics[width=.4\textwidth]{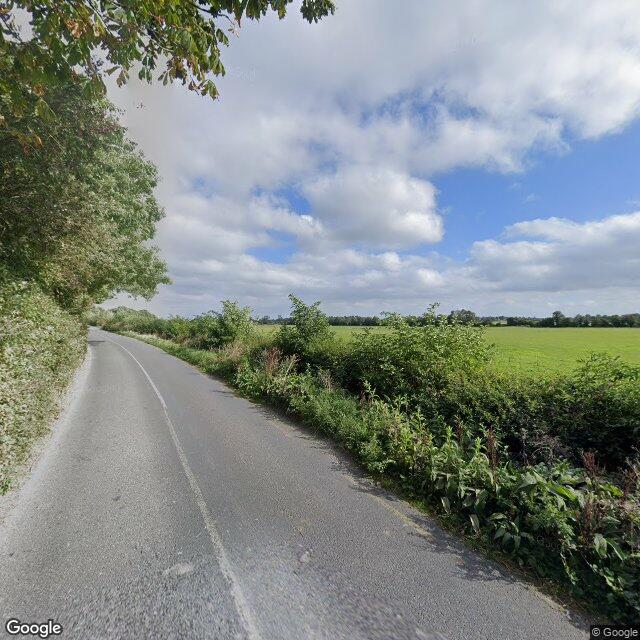}
    \includegraphics[width=.4\textwidth]{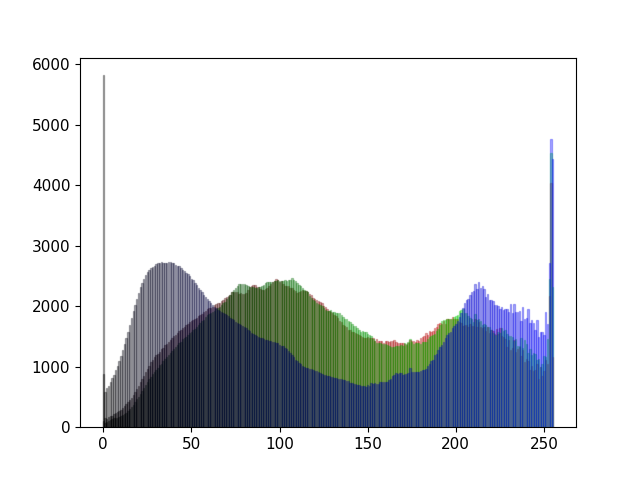}
    \caption{A Photo Taken from Google Street View in Ireland (left) and the Associated Color Histogram (right). (Left Image: Copyright by Google Inc. -- Used Under Fair Use/Academic Research)}
    \label{fig:ireland}
\end{figure}

\section{Evaluation}

A test data set was created by downloading two panoramas of randomly selected coordinates for each of the included 110 countries, resulting in a total number of 4,620 individual images. For each panorama, the system produced a probability ranking including all countries. The weightings for each individual module were optimized with another (unseen) data set beforehand, where color and captions proved to be most useful among the indicators. For the 220 guesses made, the right country landed on rank 14.7 on average, with a standard deviation of 19.1 and a median of 7. In 35 cases, the searched county was successfully ranked first.

\begin{table}
  \caption{Feature/Sub-Model Ablation Study: Each row describes a system variant with a feature or sub-model removed from the setup in the row one above}
  \label{tab:ablation-study}
\begin{tabular}{|l|r|l|l|}\hline
  System Version                               & Quality         & Evidence                    & Feature always \\
                                               & (Avg.\,Rank)    & Exploited                   & 
   Available?    \\ \hline
  Full System                                  & 14.695              & Image + Text      & n/a \\
  
  as above minus \textsl{Car License Plates}       & 14.655
              & Image (\emph{text blurred)}                       & no \\
  as above minus \textsl{Text \& Languages}    & 14.927           & Text                        & no \\
  as above minus \textsl{Object Lists}             & 15.172              & Text                        & no \\
  as above minus \textsl{Solar Position}   & 15.736              & Image  & no \\                     
  as above minus \textsl{Coloration}           & 19.727
              & Image                       & yes \\
  as above minus \textsl{Caption}           & 0
              & Text                       & yes \\ \hline
\end{tabular}
\end{table}

\section{Applications and Impact}

Identifying countries can be the first step for increasingly fine-grained
location identification. A system like our Country Guesser can serve
multiple functions:
\begin{itemize}
\item entertainment: our system  can be used as a quiz game to test one's
      ability to identify visual geographic cues about an image's
      whereabouts;
\item education: as a teaching aid to create awareness for subtle
      geographical hints of country differences; 
\item investigative journalism: to pinpoint locations of events under
      investigation by reporters and to fact-check the authenticity of potential ``fake news'' by probing for potential re-use of
      accompanying images \cite{Higgins:2021}; and
\item law enforcement: to assist crime investigations that involve
      visual evidence.
\end{itemize}
\textbf{Ethics \& Privacy Note.} Because the system makes use of data from Google 
Street~View, it inherits any potential privacy issues from it. The system by design
has a bias towards views that are visible from car-accessible roads, which includes
only portions of each country.

\section{Summary, Conclusions and Future Work}

We have presented a system demonstration comprising a country guesser that
attempts to use a range of evidence, visual features from the image itself and textual features via automatic caption generation, to compute a ranking
of most likely countries shown in an arbitrary Google Street View image.
The system's source code is available on GitLab\footnote{ \url{https://gitlab.com/Timperator/which-country-is-this} (cited 2022-01-10) }, and our
repository also contains a demo video.

\section*{Acknowledgements}
The authors wish to thank the three anonymous reviewers, whose comments improved the
quality of this paper.

%

\clearpage


\bibliographystyle{splncs04}

\bibliography{references}


\end{document}